\documentclass{article}
\usepackage{graphicx}
\PassOptionsToPackage{numbers, compress}{natbib}

\usepackage[final]{neurips_2020}
\usepackage{pdfpages}

\usepackage[utf8]{inputenc} 
\usepackage[T1]{fontenc}    
\usepackage{hyperref}       
\usepackage{url}            
\usepackage{booktabs}       
\usepackage{amsfonts}       
\usepackage{nicefrac}       
\usepackage{microtype}      
\usepackage{amsmath}
\usepackage{graphicx}
\usepackage{multirow}
\usepackage{multicol}
\usepackage{subfigure}
\usepackage{algorithm}
\usepackage{algorithmic}
\usepackage{xcolor}
\usepackage{arydshln}
\usepackage{enumitem}
\usepackage{thm-restate}
\usepackage{amsmath,amsfonts,bm,amssymb,amsthm}

\newtheorem{theorem}{Theorem}

\title{M$^3$Rec: An Offline Meta-level Model-based Reinforcement Learning Approach for Cold-Start Recommendation}

\author{%
  Yanan Wang$^{1}$, Yong Ge $^{1}$, Li Li$^{2}$, Rui Chen $^{2}$, Tong Xu $^{3}$\\
  $^1$Department of Management Information Systems, University of Arizona\\
  $^2$Samsung Research America\\
  $^3$ School of Computer Science, University of Science and Technology of China\\
  \texttt{ynwwang@email.arizona.edu}, \texttt{yongge@arizona.edu} \\
  \texttt{\{li.li1,  rui.chen1\}@samsung.com} \\
  \texttt{tongxu@ustc.edu.cn}
}

\begin{document}

\maketitle

\begin{abstract}
Reinforcement learning (RL) has shown great promise in optimizing long-term user interest in recommender systems. However, existing RL-based recommendation methods need a large number of interactions for each user to learn a robust recommendation policy. The challenge becomes more critical when recommending to new users who have a limited number of interactions. To that end, in this paper, we address the cold-start challenge in the RL-based recommender systems by proposing a meta-level model-based reinforcement learning approach for fast user adaptation. In our approach, we learn to infer each user's preference with a user context variable that enables recommendation systems to better adapt to new users with few interactions. To improve adaptation efficiency, we learn to recover the user policy and reward from only a few interactions via an inverse reinforcement learning method to assist a meta-level recommendation agent. Moreover, we model the interaction relationship between the user model and recommendation agent from an information-theoretic perspective. Empirical results show the effectiveness of the proposed method when adapting to new users with only a single interaction sequence. We further provide a theoretical analysis of the recommendation performance bound.
\end{abstract}

\section{Introduction}
Recent years have witnessed great interest in developing reinforcement learning (RL) based recommender systems~\cite{Bai2019ModelBasedRL, Chen2019GenerativeAU}, which can effectively model and optimize user's long-term interest. In the RL-based recommendation methods, the policy is learned by leveraging the collected interactions between users and recommender systems. As different users may have different interests, conventional RL-based methods need to learn a separate policy for each user, which calls for large amounts of interactions for an individual user to learn a robust recommendation policy. However, it is very difficult and expensive to obtain enough user-recommender interactions to train a robust recommendation policy. Such a challenge becomes even more critical for cold-start users who have a very limited number of interactions and prominently exist in many recommender systems. Therefore, it is crucial to learn a recommendation policy that infers the user's preference and quickly adapts to cold-start users with a limited number of interactions.


Although different users may have different interests and the interests may change over time, there are some structural similarities between these interests. \textcolor{black}{For example, users who like reading scientific books may also prefer to watch science fiction films.} Leveraging the structure similarity could be helpful for adapting recommendation policy to new users. Meta-learning~\cite{Bengio1991LearningAS, Finn2017ModelAgnosticMF} provides a solution to learn the structural similarities between user interests. However, it is still very challenging to combine meta-learning with RL-based recommender systems due to the following reasons.  First, a great amount of interactions sampled from user distributions is needed to infer user preferences. It is very difficult to get this amount of interaction for new users. Thus, it is a challenge to infer cold-start users' preferences. Second, to meta-train a recommendation policy, traditional model-free RL methods need a lot of interactions. Recent methods~\cite{Bai2019ModelBasedRL, Chen2019GenerativeAU} utilize model-based RL~\cite{Deisenroth2013ASO,Nagabandi2018NeuralND} approaches to sidestep this sample efficiency challenge by leveraging offline user data to model the environment. However, they may still need lots of offline data from each user to build the environment model, i.e., each user model. Moreover, how to better utilize the user model for policy adaptation is the third challenge. Existing RL-based recommendation methods often leverage the user model as the simulated environment to provide rewards without considering the further dependency between the user model and the recommender system.

To address the above challenges, in this paper, we propose a meta-level model-based reinforcement learning method for addressing the cold-start problem in RL-based recommendations. We introduce a user context variable to represent user preference that will be learned from the user's behavior sequences. Conditioned on the user context variable, we construct the meta-level user model and recommendation agent. For the meta-level user modeling, we recover user policy (i.e., the policy of user's decision making) and user reward function via an inverse reinforcement learning method. Then, the meta-level recommendation agent learns the policy by utilizing the meta-level user model within the framework of model-based RL. As the user model and recommendation agent interact alternately, we further propose to model the dependency between the user model and recommendation agent from an information-theoretic perspective. Specifically, we utilize mutual information between the user policy and recommendation policy to model their relation. The primary contribution of this work can be summarized as follows.
\begin{itemize}
\item We propose a novel offline meta-level model-based reinforcement learning method to address the cold-start problem in the RL-based recommender system.
\item Within our framework, we introduce a user context variable to infer user preference that enables fast adaptation on cold-start users. We also introduce a mutual information regularization within the latent policy space to capture the relation between the meta-level user model and recommendation agent, which further improves the adaption to cold-start users.
\item Empirical results compared with the state-of-the-art methods demonstrate the effectiveness of the proposed method. A theoretical analysis of the recommendation performance bound of the developed method in the offline setting is provided.
\end{itemize}

\section{Related Work}
The related work of this paper could be grouped into three categories as discussed below. 

{\bf RL-based Recommender System} There are mainly two kinds of RL methods for recommendation: model-free RL methods~\cite{chen2019top,Ie2019SlateQAT,Liebman2015DJMCAR,lu2016partially, Zhang2019TextBasedIR} and model-based RL methods~\cite{Bai2019ModelBasedRL, Chen2019GenerativeAU, Zhao2019ModelBasedRL}. Model-free RL methods assume the environment is unknown without user modeling. Model-free RL methods usually need large amounts of interactions for policy optimization.  To tackle the sample complexity challenge, model-based RL methods are applied by considering user modeling, which can predict user behavior and reward. For instance, the generative adversarial user model~\cite{Chen2019GenerativeAU} learns the user behavior model and reward function together in a unified min-max framework; then the recommendation policy is learned with reward from the trained user model. However, this model requires a large amount of data to estimate a particular user model, which is not feasible in the cold-start recommendation scenario. Besides, the user model and recommendation model are trained separately, which prevents them from benefiting from each other. ~\citeauthor{Bai2019ModelBasedRL}\cite{Bai2019ModelBasedRL} also proposed to use model-based RL for recommendation. They introduced the discriminator with adversarial training to let the user behavior and recommendation policy imitate the policy in logged offline data. The reward to train recommendation policy is weighted by the discriminator score. Their method can be seen as reward shaping~\cite{Mataric1994RewardFF,Ng1999PolicyIU}, which does not recover the true user reward function. Both of the above two methods do not properly address the cold-start challenge in the RL-based recommender system. In contrast with these methods, our method can recover the true user behavior and reward with a small amount of data by meta-learning user model and recommendation model with user context variable in a unified framework, and the mutual information regularization between the user policy and recommendation policy can benefit each other for better policy adaptation.

{\bf Meta-learning} Meta-learning aims to learn from a small amount of data and adapts quickly to new tasks ~\cite{Bengio1991LearningAS, Finn2017ModelAgnosticMF}. In context-based meta-learning~\cite{Duan2016RL2FR,Rakelly2019EfficientOM}, the approaches learn to infer the task uncertainties by taking task experiences as input. For instance, \citeauthor{Rakelly2019EfficientOM}\cite{Rakelly2019EfficientOM} proposed to learn the task context variables with probabilistic latent variables from past experiences. The model-free RL policy is trained conditioned on the task variable to improve sample efficiency. In contrast, our method learns a use context variable to infer user preference within the model-based RL framework.

{\bf Inverse Reinforcement Learning} Inverse reinforcement learning (IRL) is the problem of learning reward functions from demonstrations~\cite{Ng2000AlgorithmsFI,Abbeel2004ApprenticeshipLV, Fu2018LearningRR,Peng2019VariationalDB}, which can avoid the need for reward engineering.  For instance, \citeauthor{fu2017learning}\cite{fu2017learning} proposed an adversarial IRL (AIRL) framework to recover the true reward functions from demonstrations. IRL needs a large number of expert demonstrations to infer true reward function, which is highly expensive in the area of robotics. Recently, some works~\cite{Yu2019MetaInverseRL,Ghasemipour2019SMILeSM} try to recover reward function from a limited amount of demonstrations with the meta-IRL method by incorporating the context-based meta-learning method into AIRL framework. Comparatively, in our solution, we recover the user policy and reward function from offline user behavior data by leveraging the meta-IRL method.
To better capture the user context information into policy, We utilize a variational policy network conditioned on the user context variable. Besides, the meta-IRL learned user model serves as the environment in our meta-level model-based RL framework. 
\section{Problem Statement}\label{sec:problem_state}
In this paper, we focus on 
the cold-start problem in reinforcement learning (RL) based recommender system. Sample efficiency is a major challenge especially when the recommender system adapts to new users. To achieve a fast adaptation for new users, we investigate the cold-start problem of the RL-based recommender system from a meta-learning perspective. Let us first introduce the concept of the user context variable and then formally state the proposed meta-level model-based reinforcement learning problem that aims to address the cold-start challenge of the RL-based recommender system.

The RL-based recommendation problem is formulated as a Markov Decision Process (MDP), where the agent and environment corresponds to the recommender and user, respectively. A recommender (i.e., the learning agent) generates actions $\boldsymbol{A}_{i, t}$ (e.g., recommending $k$ items from an item set $\mathcal{D}$) at time $t$ for the $i$-th user, then the user provide feedback $x_{i,t}$ (i.e., clicked item) on the recommendation list. \textcolor{black}{The reward $r_{i,t}$ is obtained based on the clicked items} (e.g., user's purchases of clicked items or engagement time with clicked items). The recommender receives the reward and decides the next actions (e.g., another $k$ items to recommend). For each user, we have  initial state, state transition probability and reward function. We assume the distributions over users as $p(\mathcal{U})$. The context variable $\mathcal{C}$ denotes users' preferences or interests over items during their decision making process. $p(\mathcal{C}|\mathcal{U})$ denotes the distribution over context $\mathcal{C}$ for  $\mathcal{U}$.  We infer user's context variable from the user's behavior sequence. In this problem, we assume the $i$-th user's behavior sequence as $\tau_i = \{(x_{i,1}, \boldsymbol{A}_{i,1}, r_{i, 1}), (x_{i,2}, \boldsymbol{A}_{i,2}, r_{i, 2}), \cdots,
(x_{i,n}, \boldsymbol{A}_{i,n}, r_{i, n})\}$, which can be obtained in the offline user data.

To achieve fast adaptation, we infer a context variable for each user. 
Therefore, we further extend the MDP by incorporating the context variable $\mathcal{C}$. Different environments will induce different user context variables $c \in \mathcal{C}$. In the following notations, we omit the script of user index, which is reflected in the user context variable $c$.  The state of the environment is denoted by $\boldsymbol{s}_{t} \in \mathcal{S}$, which corresponds to one user's historical clicked item sequence before time $t$. We assume the users follow their policies when making choice, which is referred as user policy $\pi_{\phi}(x_t|\boldsymbol{s}_{t},  \boldsymbol{A}_{t}, c)$. This user policy indicates that user clicks item $
x_t$ chosen from item set $\boldsymbol{A}_{t}$ based on state $s_t$ with a specific user context variable $c$. Then, 
state transition function $\mathcal{P}: \mathcal{C} \times \mathcal{S} \times \mathcal{A} \times \mathcal{S} \rightarrow \mathbb{R}$ can be modeled by user policy $\pi_{\phi}$, which represents the probability of transferring to state $\boldsymbol{s}_{t + 1}$ given state $\boldsymbol{s}_{t}$, recommendation list $\boldsymbol{A}_{t}$ and the specific user context variable $c$.
User's reward function is denoted as  $r_w(\boldsymbol{s}_{t}, x_{t}, \boldsymbol{A}_{t}, c)$, which represents the reward after the user chooses item $x_t$ from $\boldsymbol{A}_{t}$ in state $\boldsymbol{s}_{t}$ with context variable $c$. For the recommender agent, the policy is defined as $\pi_{\theta}(\boldsymbol{A}_{t}|\boldsymbol{s}_{t}, c)$, where the recommender decides recommendation list $\boldsymbol{A}_{t}$ from item set $\mathcal{D}$ in state $\boldsymbol{s}_{t}$ conditioned on user context variable $c$. In this problem, the user reward and recommender reward are assumed to be the same.

Now, we formally define our problem. Given a set of users sampled from user distribution $p(\mathcal{U})$,
we meta-learn a user model, which recovers the user policy and user reward from offline data. Assisted by the meta-level user model $\pi_{\phi}$, we meta-train a recommendation policy $\pi_{\theta}$ to adapt to users conditioned on user context variable. During the meta-test stage, the recommendation policy adapts to new users conditioned on only a single user behavior sequence.

\section{Method}
In this section, we present the proposed Mutual information regularized Meta-level Model-based reinforcement learning approach for cold-start Recommendation (M$^3$Rec).
\subsection{Mutual Information Regularized Meta-level Model-based Reinforcement Learning}
We tackle the cold-start problem of RL-based recommendation from a meta-learning perspective. We propose to condition the user model and recommendation agent on the user context variable, which can derive a context-aware user and recommendation policy as well as a user reward function.
To improve the meta-training efficiency, we further model the dependency between user policy and recommendation policy using mutual information regularization.
\paragraph{Meta-level User Model as Inverse Reinforcement Learning}
As the conditioned user policy and reward function are unknown, we aim to recover both the user policy and reward function from offline data in the meta-level user model. 

As the user context variable $\mathcal{C}$ is sampled from distribution $p(\mathcal{C}|\mathcal{U})$, it can be estimated from the offline user's behavior sequence $\tau$. We infer the $i$-th user's context variable $c_i$ as $p_f(c_i|\tau_i)$. To learn a conditioned user policy $\pi_{\phi}(x_t|\boldsymbol{s}_{t},  \boldsymbol{A}_{t}, c)$, it must encode the salient information of the user context variable $c$ into user policy representation. 
\textcolor{black}{To make the user policy better aware of the user context variable, we adopt the variational inference approach~\cite{Kingma2014AutoEncodingVB, Sohn2015LearningSO} to infer the latent user policy variable $z_u$ with a  user context variable.} It can be generated from variational distribution $q_{\phi}(z_u|\boldsymbol{s}_{t}, c)$. To optimize the parameters for learning $z_u$, we maximize the lower bound of $\log p_{\phi}(\boldsymbol{s}_{t} | c)$:
\begin{equation}
    \log p_{\phi}(\boldsymbol{s}_{t} | c) \geq
    \mathbb{E}_{q_{\phi}(z_u|\boldsymbol{s}_{t}, c)}[\log p_{\phi}(\boldsymbol{s}_{t}|z_u,  c)] - \beta D_{\mathrm{KL}}(q_{\phi}(z_u|\boldsymbol{s}_{t}, c) \| p_{\phi}(z_{u} | c)),
\end{equation}
where the first term is optimized to reconstruct current state $\boldsymbol{s}_t$. Empirically, we found that training the decoder $p_{\phi}(\boldsymbol{s}_{t}|z_u,  c)$ to predict the next state $\boldsymbol{s}_{t+1}$ performs better, which models the user's dynamics. The second term constrains the latent policy variable with a Gaussian prior. Then, the user's choice for item $x_t$ can be obtained as $\pi_{\phi}(x_t|z_u, \boldsymbol{
A}_t)$. 

To recover both the actual user policy and  reward function, we utilize offline user data $\tau$.
Inspired by Adversarial Inverse Reinforcement Learning (AIRL)~\cite{fu2017learning}, we recover both the conditioned variational user policy and conditioned reward function from offline data by optimizing the following objective:
\begin{equation*}
    \min _{\pi_{\phi}} \max _{D_{\omega}} \mathbb{E}_{p(\boldsymbol{A},c)}[\mathbb{E}_{\rho^{true }(\boldsymbol{s}, x | \boldsymbol{A}, c)}[\log D_{\omega}(\boldsymbol{s}, x, \boldsymbol{A}, c)]+\mathbb{E}_{\rho^{\pi_{\phi}}(\boldsymbol{s}, x | \boldsymbol{A}, c)}[\log (1-D_{\omega}(\boldsymbol{s}, x, \boldsymbol{A}, c))]],
\end{equation*}
where the discriminator function is $D_{\omega}(\boldsymbol{s}, x,\boldsymbol{A},c)=\exp (g_{\omega}(\boldsymbol{s}, x, \boldsymbol{A},c)) /(\exp (g_{\omega}(\boldsymbol{s}, x, \boldsymbol{A},c))+\pi_{\phi}(x | \boldsymbol{s}. \boldsymbol{A},c))\nonumber$. $g_{\omega}$ contains the reward approximator $r_{\omega}$ and the reward shaping term $h_{\varphi}$:
$g_{\omega}(\boldsymbol{s}, x, \boldsymbol{A},c) = r_{\omega}(\boldsymbol{s}, x, \boldsymbol{A},c) + \gamma h_{\varphi}(\boldsymbol{s}^{\prime}, c) - h_{\varphi}(\boldsymbol{s}, c)$, where $\gamma$ is the discount factor and $\boldsymbol{s}^{\prime}$ is the next state of state $\boldsymbol{s}$. 
To infer a user's context variable $c$, we first sample a user $u$ from distribution $p(\mathcal{U})$. Then $c$ can be inferred using sampled true user behavior sequence $\tau_{u}^c$ by $p_f(c|\tau_{u}^c)$ \textcolor{black}{as aforementioned}. The tuple of state $\boldsymbol{s}$, user choice $c$ and recommendation list $\boldsymbol{A}$ from $\rho^{true }$ is sampled from true user behavior sequence $\tau_{u}^{true}$. $\{\boldsymbol{s}, x, \boldsymbol{A}\}$ from $\rho^{\pi_{\phi}}$ can be sampled from rollouts $\tau_u$ generated by policy $\pi_{\phi}$ conditioned on user context variable $c$.

Specifically, during training, We can alternately update parameters of user policy $\pi_{\phi}$ and discriminator $D_w$. The objective for training the user policy $\pi_{\phi}$ is:
\begin{equation}\label{eq:user_policy}
    \max _{\phi} \mathbb{E}_{u \sim p(\mathcal{U}), c \sim p_f(c|\tau_u^{c}),\tau_u \sim \rho^{\pi_{\phi}}(\tau_u | \boldsymbol{A}, c)} \sum_{t=1}^{T} \log (D_{w}(\boldsymbol{s}_t, x_t, \boldsymbol{A}_t, c) - \log (1-D_{w}(\boldsymbol{s}_t, x_t, \boldsymbol{A}_t, c)),
\end{equation}
where we can train the meta-level user policy $\pi_{\phi}$ using policy gradient algorithm.

The objective for training the discriminator is:
\begin{align}\label{eq:discriminator}
\max _{D_{\omega}} \mathbb{E}_{u \sim p(\mathcal{U})} [&\mathbb{E}_{c \sim p_f(c|\tau_u^{c}), \tau_u \sim \rho^{\pi_{\phi}}(\tau_u | \boldsymbol{A}, c)} \sum_{t=1}^{T}\log  (1-D_{w}(\boldsymbol{s_t}, x_t, \boldsymbol{A_t},c) +\nonumber\\
&\mathbb{E}_{c \sim p_f(c|\tau_u^{c}), \tau_u^{true} \sim \rho^{true}(\tau_u^{true})} \sum_{t=1}^{T}\log  D_{w}(\boldsymbol{s_t}, x_t, \boldsymbol{A_t},c)].
\end{align}
Similar to AIRL in ~\cite{fu2017learning}, when training the conditioned variational user policy and conditioned discriminator to optimality, we can recover the true user policy and the true reward function up to a constant, which approximate the real user model. \textcolor{black}{Therefore, we can also utilize the offline data to estimate the meta-level user model by maximizing the likelihood of $\log \pi_{\phi}(x_t|\boldsymbol{s}_t,  \boldsymbol{A}_t, c)$ sampling offline data $\tau_u^{true} \sim \rho^{true}(\tau_u^{true})$}.
\paragraph{Meta-level Recommendation Agent}
The recommendation agent aims to maximize the cumulative user reward and adapt to new users.
To facilitate fast adaptation, we condition the recommendation policy $\pi_{\theta}(\boldsymbol{A}_{t}|\boldsymbol{s}_{t}, c)$ on the user context variable $c$. Similar to the meta-level user model, we use variational recommendation policy conditioned on the user context variable to enable the recommendation policy \textcolor{black}{to be aware of the user preference. The latent recommendation policy variable is denoted as $z_{rec}$ induced from variational distribution $q_{\theta}(z_{rec}|\boldsymbol{s}_t, c)$ . We optimize the lower bound of $p_{\theta}(\boldsymbol{s}_t | c)$:}
\begin{equation}
    \log p_{\theta}(\boldsymbol{s}_{t} | c) \geq
    \mathbb{E}_{q_{\theta}(z_{rec}|\boldsymbol{s}_{t}, c)}[\log p_{\theta}(\boldsymbol{s}_{t}|z_{rec},  c)] - \beta D_{\mathrm{KL}}(q_{\theta}(z_{rec}|\boldsymbol{s}_{t}, c) \| p_{\theta}(z_{rec} | c)).
\end{equation}
Then, based on the latent recommendation policy variable, the agent will generate a recommendation list of size $k$ under policy $\pi(\boldsymbol{A}|z_{rec})$. Specifically, the probability that the user's clicked item $x_t \in \boldsymbol{A}_t$  is $\pi(x_t|z_{rec})$. The objective for the recommendation policy is as follows:
\begin{equation}\label{eq:rec_policy}
    \max _{\theta} \mathbb{E}_{u \sim p(\mathcal{U}), c \sim p_f(c|\tau_u^{c}), \tau_u \sim \rho^{\pi_{\theta}}(\tau_u | \boldsymbol{A}, c)} \sum_{t=1}^{T} r_w(\boldsymbol{s}_t, x_t, \boldsymbol{A}_t, c),
\end{equation}
which can be optimized by using policy gradient algorithm.

\textcolor{black}{As shown above, when training the meta-user model to optimality, the reward function $r_w(\boldsymbol{s}, x, \boldsymbol{A}, c)$ will recover the true reward to a constant, which reduces bias. Therefore, $\pi_{\theta}(\boldsymbol{A}_{t}|\boldsymbol{s}_{t}, c)$ can also be optimized by policy gradient algorithm using the
true offline data $\tau_u^{true} \sim \rho^{true}(\tau_u^{true})$}.
\paragraph{Mutual Information Regularization for Policy Adaptation}
As shown in the setting of RL-based recommendation system, the user model and recommendation agent interact alternately. Therefore, there is a high dependency between the user policy and recommendation policy. It is necessary to establish influence function between these two policies. From the information-theoretic perspective, the mutual information between the latent user policy variable $z_{u}$ and the latent recommendation policy variable $z_{rec}$ can measure the influence relationship. We further analyze the influence of mutual information regularization in Sec~\ref{sec:theoretical_analysis}.

The mutual information between two variables is defined as:
\begin{equation}
\mathcal{I}(z_{u} ; z_{rec})=D_{\mathrm{KL}}\left(\mathbb{P}_{z_{u} z_{rec}} \| \mathbb{P}_{z_{u}} \otimes \mathbb{P}_{z_{rec}}\right),
\end{equation}
where $\mathbb{P}_{z_{u} z_{rec}}$ is the joint distribution. $\mathbb{P}_{z_{u}}$ and $\mathbb{P}_{z_{rec}}$ are marginal distributions.

we want to maximize $\mathcal{I}(z_{u} ; z_{rec})$ , which can assist the adaptation of the recommendation policy as well as the estimation of user policy. However, mutual information is difficult to estimate in high dimension space. Inspired by ~\cite{belghazi2018mine, nowozin2016f}, we maximize the lower bound of Jensen-Shannon mutual information  $\mathcal{I}^{(JSD)}(z_{u}; z_{rec})$ for stable training as:
\begin{align}\label{mi_bound}
\mathcal{I}^{(JSD)}(z_{u}; z_{rec}) \geq \sup _{\psi \in \Psi} \mathbb{E}_{\mathbb{P}_{z_{u} z_{rec}}}\left[-s p\left(-T_{\psi}(z_{u}, z_{rec})\right)\right]
-\mathbb{E}_{\mathbb{P}_{z_{u}} \otimes \mathbb{P}_{z_{rec}}}\left[\operatorname{sp}\left(T_{\psi}(z_{u}, z_{rec})\right)\right],
\end{align}
where $T_{\psi}: \mathcal{X} \times \mathcal{Y} \rightarrow \mathbb{R}$ is a
neural network function with parameter $\psi$ and $\operatorname{sp}(z)=\log (1+e^{z})$ is the softplus function.

During training, we maximize the lower bound in Eq.~\ref{mi_bound} to update corresponding parameters in the meta-level user model and meta-level recommender agent model.
\vspace{-0.3cm}
\subsection{Theoretical Analysis}\label{sec:theoretical_analysis}
In this section, we provide a theoretical analysis of the performance bound of the recommender policy when adapting to meta-test users by our meta-level model-based RL framework.

To provide our theoretical analysis, let us first introduce some notations. Here, we slightly abuse the notation for simplicity. We denote action at time $t$ as $a_t$, which corresponds to $x_t$ and $\boldsymbol{A}_t$ defined in Section~\ref{sec:problem_state} for user and recommender agent respectively. We use $\mu_{u_i}^{\pi_{\theta}}=\frac{1}{T} \sum_{t=0}^{T} P(\boldsymbol{s}_{t}=\boldsymbol{s}, a_{t}=a)$ to denote the average state action ($\boldsymbol{s}$, $a$) visitation distribution when executing recommendation policy $\pi_{\theta}$ in user model $u_i$, where $u_i$ can be the approximated user model $u_i^{m}$ in our meta-level model-based RL or the true user model $u_i^{w}$ in the real world. Then for user $u_i$, there is modeling error between $u_i^{m}$ and $u_i^{w}$ under state-action distribution $\mu$:
$\ell(u_i^{m}, \mu)=\mathbb{E}_{(\boldsymbol{s}, a) \sim \mu}[D_{KL}(P_{u_i^{w}}(\cdot | \boldsymbol{s}, a), P_{u_i^{m}}(\cdot | \boldsymbol{s}, a))]$, where $D_{KL}$ is the KL divergence. $P_{u_i^{w}}$ and $P_{u_i^{m}}$ represents user transition dynamics (i.e., user policy $\pi_{\phi}$) in true user model $u_i^{w}$ and approximated meta-level user model $u_i^{m}$ respectively. The performance of recommendation policy $\pi_{\theta}$ under user model $u_i$ is $J(\pi_{
\theta}, u_i)=\mathbb{E}[\sum_{t=0}^{\infty} \gamma^{t} r_{u_i, t}]$, where $\gamma$ is the discount factor. Then, we can get the recommendation policy performance bound learned in our meta-level model-based RL framework. 

\begin{theorem}\label{sec:lemma1}
Suppose the meta-level user model and meta-level recommendation policy is trained to optimality on meta-training users.
When adapting to a meta-test user with few behavior sequences to infer the user context variable, the test user's policy is obtained as $\pi_{\phi}(x|\boldsymbol{s}, \boldsymbol{A}, c_{test})$ with the corresponding recommendation policy as $\pi_{\theta}(\boldsymbol{A}|\boldsymbol{s}, c_{test})$. Suppose the modeling error of this test user model $\ell(u_{test}^m, \mu_{u_{test}^w}^{\pi_{\theta}}) \leq \epsilon_{u_{test}^m}^{adapt}$. The performance of  $\pi_{\theta}(\boldsymbol{A}|\boldsymbol{s}, c_{test})$ under this test user's model satisfies $J(\pi_{\theta}, u_{test}^m) \geq \sup _{\pi_{\theta}^{\prime}} J(\pi_{\theta}^{\prime}, u_{test}^m)-\epsilon_{\pi_{\theta}}^{adapt}$.
Let $\pi_{\theta}^*$ denotes the optimal recommendation policy and the corresponding performance is $J_{u_{test}^w}^{*}=\sup _{\pi_{\theta}^{\prime}} J(\pi_{\theta}^{\prime}, u_{test}^w)$. We also suppose the modeling error of this test user model on the optimal recommendation policy $\pi_{\theta}^{\prime}$ is $\ell(u_{test}^m, \mu_{u_{test}^w}^{\pi_{\theta}^*}) \leq \epsilon_{u_{test}^m}^{adapt}$ as the meta-training process can help the model adapt to different recommendation policy. Then, the performance bound between the learned recommendation policy $\pi_{\theta}$ and the optimal policy $\pi_{\theta}^*$ on real meta-test users is as follows:
\begin{align*}
    J\left(\pi_{\theta}^*, u_{t e s t}^{w}\right)- J\left(\pi_{\theta}, u_{t e s t}^{w}\right) &\leq \epsilon_{\pi_{\theta}}^{adapt} + \frac{4 \gamma R_{\max }\sqrt{\epsilon_{u_{test}^m}^{adapt}}}{(1-\gamma)^{2}}
\end{align*}
\end{theorem}

\textbf{Remark.} In this theorem, the gap between the recommender policy performance in our model trained on meta-training users and the optimal policy in the real-world test users comes from two error terms.

The first term is related to the sub-optimality of the meta-policy optimization as well as the generalization error of meta-level recommendation policy to the meta-test user.
It can be reduced by the sufficient training of meta-level recommendation policy. The mutual information regularization between user policy and recommendation policy can also help reduce error $\epsilon_{\pi_{\theta}}^{adapt}$. 

The second term is related to the user model adaptation error $\epsilon_{u_{test}^m}^{adapt}$ on the new meta-test user recommendation policy $\pi_{\theta}$ and its optimal recommendation policy $\pi_{\theta}^*$. As our meta-level user model learns from a distribution of users and optimize its prediction performance on different meta-level recommendation policies, the model adaptation error $\epsilon_{u_{test}^m}^{adapt}$ can be small. Intuitively, the mutual information regularization between meta-level user model and meta-level recommendation agent can reduce the uncertainly of visiting out-of-distribution stat-action, which further reduce the model estimation error in the offline setting.

Compared with recent offline reinforcement methods built on uncertainty estimation with soft or hard constraints~\cite{yu2020mopo, kidambi2020morel}, we utilize the meta-learning and mutual information regularization between model and policy to reduce the model estimation error. Specifically, the meta-learning helps the model to accurately simulate on a wide range of policies and the mutual information regularization provides implicit constraint between model and policy to avoid visiting out-of-
distribution state-action too much in the offline setting.
\section{Experiment}
In this section, we carry out the experiments in two settings: simulated online evaluation and offline evaluation with a real-world dataset. The proposed method is compared with state-of-the-art baselines to demonstrate the effectiveness of our solution in the cold-start recommendation problem. We denote the proposed mutual-information regularized meta-level model-based reinforcement learning method as M$^3$Rec. Specifically, the 
selected state-of-the-art baselines are \textbf{Meta-LSTM}, \textbf{Meta-Policy Gradient (Meta-PG)}, \textbf{IRecGAN}~\cite{Bai2019ModelBasedRL}, \textbf{Generative Adversarial User Model (GAN-PG)}~\cite{Chen2019GenerativeAU}, \textbf{MeLU}~\cite{lee2019melu}.

For all the RL-based methods, REINFORCE~\cite{Williams1992SimpleSG} algorithm is used for policy optimization. \textcolor{black}{During meta-test stage, we utilize a single interaction sequence for each test user, which is similar to the one-shot learning setting~\cite{Bruce2017OneShotRL, Duan2017OneShotIL}.}
\subsection{Simulated Online Evaluation}
As it is difficult to conduct the online evaluation by interacting with real users, we carry out the online evaluation in a simulated environment by following previous works~\cite{Bai2019ModelBasedRL, Chen2019GenerativeAU}.

To simulate the behavior of different users, we utilized an open-sourced simulator\footnote{\url{https://github.com/google-research/recsim}}~\cite{Ie2019RecSimAC} for recommendation system, which provides sequential interaction with users. We first generate offline data for model training. The offline data contains 2,500 users, where we configure the simulator with different user parameters. After training the model using the offline data, we apply the model in the simulator by interacting with users to test the online performance. We configure another 500 users as the test user set, whose parameter configurations are different from users used in the offline data.  Table~\ref{sec:online_results} shows the average rewards of all competing methods. It can be observed that the proposed method M$^3$Rec outperforms the baseline methods, especially the two model-based RL methods without meta-learning. By inferring a user context variable, our method can effectively adapt to the new users.

\begin{table}[t]
\caption{Online evaluation results of average reward with different recommendation list sizes $k$.}\label{sec:online_results}
\small
\begin{tabular}{l|cccccc}
\hline
Model     & Meta-LSTM & Meta-PG & IRecGAN & GAN-PG & MeLU &M$^3$Rec (Ours) \\ \hline
$k=3$  &$38.86_{\pm 0.42}$           & $35.43_{\pm 0.33}$         &$37.53_{\pm 0.61}$         &$36.46_{\pm 0.54}$ &$36.54_{\pm 0.47}$        &$\textbf{40.72}_{\pm 0.66}$                         \\
$k=5$  &$39.01_{\pm 0.61}$           &$35.54_{\pm 0.54}$         &$37.76_{\pm 0.91}$         &$36.74_{\pm 0.94}$ & $36.64_{\pm 0.99}$        &$\textbf{40.80}_{\pm 0.50}$                         \\ \hline
\end{tabular}
\end{table}

\subsection{Offline Evaluation With Real-world Dataset}
\begin{table}[ht]
\caption{Offline evaluation results.}\label{sec:real_data_result}
\begin{small}
\scalebox{1.0}{\begin{tabular}{l|cccccc}
\hline
Model     & Meta-LSTM & Meta-PG & IRecGAN & GAN-PG & MeLU & M$^3$Rec (Ours) \\ 
\hline 
P@1 (\%)  &$4.43_{\pm 1.00}$           &$5.99 _{\pm 0.18}$         &$6.83_{\pm 1.29}$         &$3.62_{\pm 0.49}$  & $5.47_{\pm 0.33}$     &$\textbf{7.26}_{\pm 0.51}$                         \\ 
P@5 (\%)  &$22.81_{\pm 1.59}$           &$20.85 _{\pm 0.98}$         &$24.36_{\pm 1.45}$         &$17.26_{\pm 1.81}$  & $21.73_{\pm 0.92}$     &$\textbf{28.42}_{\pm 0.32}$                         \\ 
P@10 (\%) &$36.81_{\pm 1.43}$           &$32.13_{\pm 1.17}$         &$36.84_{\pm 1.18}$         &$29.09_{\pm 1.30}$  &  $34.27_{\pm 0.53}$    &$\textbf{40.02}_{\pm 0.34}$                   
\\
NDCG@5 (\%)  &$13.75_{\pm 1.25}$           &$13.61_{\pm 0.70}$         &$15.60_{\pm 1.39}$         &$10.42_{\pm 0.96}$  & $13.81_{\pm 0.38}$     &$\textbf{17.89}_{\pm 0.09}$                       \\ 
NDCG@10 (\%)  &$18.24_{\pm 1.20}$           &$17.21_{\pm 0.74}$         &$19.60_{\pm 1.32}$         &$14.19_{\pm 0.87}$  & $17.80_{\pm 0.24}$     &$\textbf{21.63}_{\pm 0.24}$

\\
Recall@5 (\%)  &$5.75_{\pm 0.66}$           &$4.98 _{\pm 0.52}$         &$5.43_{\pm 0.71}$         &$4.57_{\pm 1.09}$  & $5.94_{\pm 0.71}$     &$\textbf{7.04}_{\pm 0.15}$
\\ 
Recall@10 (\%)  &$8.69_{\pm 0.52}$           &$8.02_{\pm 0.22}$         &$8.49_{\pm 0.32}$         &$7.78_{\pm 1.02}$  & $8.97_{\pm 0.28}$     &$\textbf{9.16}_{\pm 0.24}$    
\\
\hline
\end{tabular}}
\end{small}
\end{table}
We further validate the effectiveness of our proposed method with a real-world recommendation dataset. We utilize the dataset from CIKM CUP 2016 for offline reranking.
The performance of offline evaluation is shown in Table~\ref{sec:real_data_result}, where the Precision, Normalized Discounted Cumulative Gain (NDCG), Recall metrics are utilized. It can be clearly observed that out proposed method significantly outperforms all the baseline methods, which demonstrates the effectiveness of our method for the cold-start recommendation problem.

\vspace{-0.3cm}
\section{Conclusion}
\vspace{-0.2cm}
In this paper, we proposed a novel offline meta-level model-based reinforcement learning method to address the cold-start problem in RL-based recommendations. In the developed method, we introduced a user context variable for inferring user preference and a mutual information regularization for capturing the dependency between the proposed meta-level user model and meta-level recommendation agent, both of which together enable fast adaption to cold-start users. To improve the adaption efficiency, we proposed to recover the user policy and reward via an inverse reinforcement learning approach. In addition to both online and offline evaluations that demonstrate the effectiveness of our approach, we provided a theoretical analysis of the recommendation performance bound of the developed method. Although we test the proposed offline RL method on the recommendation task, it is a general offline RL framework and can be applied in other applications like robot control.
\bibliographystyle{plainnat}
\bibliography{neurips_2020}
\end{document}